\documentclass[runningheads,a4paper]{llncs}

\usepackage{graphicx}
\usepackage{times}
\usepackage{amsfonts,amssymb}
\usepackage{amsmath}
\usepackage{url}

\newcommand{\keywords}[1]{\par\addvspace\baselineskip
\noindent\keywordname\enspace\ignorespaces#1}

\begin{document}

\mainmatter  

\title{Banzhaf Random Forests}

\titlerunning{Lecture Notes in Computer Science: Authors' Instructions}

%
%
\author{Jianyuan Sun\thanks{PhD Student}$^\dag$%
\and Guoqiang Zhong$^\dag$\thanks{$\dag$ Same contribution}\thanks{To whom correspondence should be addressed.} \and Junyu Dong \and Yajuan Cai}
\authorrunning{J. Sun, G. Zhong, Y. Cai and J. Dong}

\institute{Department of Computer Science and Technology,\\
Ocean University of China, Qingdao 266100, China\\
sunjianyuan11@163.com, gqzhong@ouc.edu.cn, cyj-ouc@hotmail.com, dongjunyu@ouc.edu.cn}

%
%

\toctitle{Lecture Notes in Computer Science}
\tocauthor{Authors' Instructions}
\maketitle

\begin{abstract}
 Random forests are a type of ensemble method which makes predictions by combining the results of several \emph{independent} trees. However, the theory of random forests has long been outpaced by their application. In this paper, we propose a novel random forests algorithm based on cooperative game theory. \emph{Banzhaf power index} is employed to evaluate the power of each feature by traversing possible feature coalitions. Unlike the previously used information gain rate of information theory, which simply chooses the most informative feature, the Banzhaf power index can be considered as a metric of the importance of each feature on the \emph{dependency} among a group of features. More importantly, we have proved the consistency of the proposed algorithm, named Banzhaf random forests (BRF). This theoretical analysis takes a step towards narrowing the gap between the theory and practice of random forests for classification problems. Experiments on several UCI benchmark data sets show that BRF is competitive with state-of-the-art classifiers and dramatically outperforms previous consistent random forests. Particularly, it is much more efficient than previous consistent random forests.
\keywords{random forests, Banzhaf power index, cooperative game, classification}
\end{abstract}

\section{Introduction}

Ensemble methods are learning algorithms that construct a set of classifiers and combine them to classify new unseen data~\cite{zhou2012ensemble}. Random forests are a type of ensemble method based on combination of several independent decision trees~\cite{randfore}. In recent years, the random forests framework and its variants have been successfully applied in practice as a general classification and regression tool. Particularly, random forests have been widely used in computer vision~\cite{lepetit2006keypoint},~\cite{ozuysal2007fast},~\cite{shotton2013real},~\cite{zikic2013atlas} and pattern recognition applications~\cite{winn2006object},~\cite{yin2007tree},~\cite{bosch2007image},~\cite{shotton2008semantic}, which promotes the state-of-the-art in performance. Despite their successful applications, the theoretical analysis of random forest models is still very difficult, even the basic mathematical properties are very hard to understood. In~\cite{biau2008consistency} and~\cite{biau2012analysis}, Biau and colleagues tries to narrow the gap between the theory and practice of random forest. However, the proposed models in these two papers cannot deliver effective results and their running is not efficient.

In this paper, we introduce a novel random forests algorithm based on the cooperative game theory. We adopt the Banzhaf power index to evaluate the power of each feature by traversing all possible coalitions. Due to this, we call the proposed algorithm Banzhaf random forests (BRF). Different from the previously used information gain rate of information theory, which simply chooses the most informative feature, the Banzhaf power index measures the importance of each feature on the dependency among a group of features (coalition). More importantly, We reasonable proved the consistency of the forest, it has made a contribution to narrow the theory and practice gap for random classification forests problems.

The rest of this paper is organized as follows. In Section 2, we provide a brief overview of existing random forests models and analyze their advantage and disadvantage. In Section 3, we introduce the general random forests framework, including the construction of trees and randomness injection. Section 4 describes the proposed algorithm, Banzhaf random forests (BRF), in detail, while Section 5 is devoted to the justification of the consistency of BRF. Section 6 shows the experimental results on some UCI benchmark data sets and Section 7 concludes this paper.

\section{Related work}

Classic random forests introduced by Breiman~\cite{randfore} combine several decision trees~\cite{breiman1984classification} with bagging~\cite{breiman1996bagging}. The main idea of random forests is based on the early work of~\cite{ho1998random} on the random subspace method, the feature selection work of~\cite{amit1997shape}, the way of random split selection of~\cite{dietterich2000experimental}. Based on the seminal work of Breiman ~\cite{randfore}, ~\cite{kwok2013multiple} suggests that it is best to average across sets of trees with different structures but not any of the constituent trees. Criminisi et al. ~\cite{criminisi2012decision} present a unified, efficient model of random decision forests which can be applied to a number of machine learning, computer vision and medical image analysis tasks. With the development of random forests in recent years, they have been applied to a wide variety of real world problems~\cite{svetnik2003random},~\cite{prasad2006newer},~\cite{cutler2007random},~\cite{criminisi2013decision}.

Despite the successful applications of random forests in practice, the mathematical properties behind them have not been well understood. For example, the early theoretical work of~\cite{breiman2004consistency}, which is essentially based on mathematical heuristics, is not formalized to rigorous theory.

In theory, there are two main properties of theoretical interests related to random forests. One is the consistency of the models, that whether it can converge to an optimal solution as the data set grows infinitely large. The other is the rate of convergence. Our paper mainly focuses on consistency, which~\cite{biau2008consistency} has proved that Breiman's random forests cannot guarantee.

To design consistent random forests, many researchers have struggled in this trend. Meinshausen~\cite{meinshausen2006quantile} has shown that an algorithm of random forests for quantile regression is consistent; Ishwaran and Kogalur~\cite{ishwaran2010consistency} have shown the consistency of their survival forests model; Denil et al.~\cite{denil2013consistency} show the consistency of an online version of random forests, while~\cite{denil2013narrowing} presents a new random regression forests. These consistent models can be applied to either regression, survival or online settings, but not to batch classification settings where all the training data can be used together for learning. In this paper, we propose a novel random forests model based on the cooperative game theory for multi-class classification problems. The consistency of the proposed algorithm is also proved.

Two more closely related papers to our work are~\cite{biau2008consistency} and~\cite{biau2012analysis}. \cite{biau2008consistency} proves the consistency of some popular averaging classifiers, including random forests. Specifically, the authors take~\cite{randfore} as a weighted layered nearest neighbor classifier from the perspective of taxonomy proposed by~\cite{lin2006random}. Unfortunately, this property prevents the consistency of random tree classifiers. To remedy the inconsistency of tree classifiers, the authors suggest the technique introduced in~\cite{gyorfi1996probabilistic}. Moreover, \cite{biau2008consistency} has also proposed a scale-invariant version of random forests with consistency. Recently, \cite{biau2012analysis} presents a new model of random forests, which is similar to the original algorithm of~\cite{randfore}. The main difference between these two models is in how random features are selected. \cite{biau2012analysis} requires a second independent data set to evaluate the importance index of each feature and uses this property to prove the consistency for their algorithm, while the model of~\cite{randfore} doesn't need the second data set. In this paper, we use the Banzhaf power index to evaluate the power of each feature by traversing all possible feature coalitions, but not employing the second data set. The consistency of the proposed algorithm is theoretically guaranteed.

\section{Random Forests}

In this section we briefly review the random forests framework. Typically, random forests are built by combining the predictions of several trees, each of which is trained in isolation. Unlike in boosting~\cite{schapire2013boosting}, where the base models are trained and combined using a dynamic weighting scheme, the trees are trained independently and the predictions of the trees are combined through averaging or majority voting. For a more comprehensive review, please refer to~\cite{randfore} and~\cite{criminisi2012decision}.

To construct a random tree, three core steps are required: the first is the method for splitting the tree nodes; the second is the type of predictor to use in each leaf, and the third is the method of injecting randomness into the trees.

In a typical method for splitting nodes, splitting depends on whether or not they exceed a threshold value in a chosen feature. Alternatively, for linear splits, a linear combination of features are compared with a threshold to make decision. The threshold value in either case can be chosen randomly or by optimizing a function of the data. For example, the Gini index and information gain rate are commonly used. In this paper, we choose the midpoint of a feature as the splitting threshold, which leads to the proposed algorithm to be very efficient, especially in the case of large scale applications.

In order to split a node of each tree, candidate features of data are generated and a criterion is evaluated to choose between them. A simple strategy, as in the models analyzed in~\cite{biau2008consistency}, is to choose among the features uniformly at random. A more common approach is to choose the candidate split which optimizes a purity function over the nodes that would be created. Particularly, two typical choices are to maximize the information gain~\cite{hastie2009elements} and minimize the Gini index. In our Banzhaf random forests, we use the Banzhaf power index of the cooperative game theory~\cite{banzhaf1964weighted}, which measures the distribution of power among the features on the data sets.

For the choice of predictors,~\cite{criminisi2012decision} propose several different leaf predictors for regression and other tasks. One common consideration is to average predictors over the training points which fall in that leaf. The other consideration may based on majority voting with points in that leaf. In our work, we take the last strategy.

It is important to inject randomness into the trees for random forests. This can be achieved in several ways. One choice is on the features to be split at each node; the other one is the coefficients for random combinations of features. One common method is to build each tree using a bootstrapped or sub-sampled data set. In this way, each tree in the forest is trained on slightly different data, which introduces differences between the trees. Similar to ~\cite{randfore}, our work uses a bootstrapped method to inject randomness into each tree.

\section{Banzhaf Random Forests}

In this section, we describe the proposed algorithm, Banzhaf random forest (BRF), in detail. Firstly, we introduce some basic concepts of cooperative game theory. Secondly, based on the Banzhaf power index, we introduce the way to construct the randomized trees. Thirdly, we combine the Banzhaf trees to formulate the Banzhaf random forests. Finally, we present the prediction method about the Banzhaf random forests.

\subsection{Basic concepts of cooperative game theory}

Cooperative game theory mainly studies an `acceptable' way of distributing gains collectively achieved by a group of cooperating agents~\cite{chalkiadakis2011computational}. A cooperative profit game $\Gamma=(\mathcal{N},\gamma)$ consists of a player set $\mathcal{N}=\{1,2,...,n\}$ and a characteristic function $\gamma:2^\mathcal{N}\rightarrow R$. For each subset $\mathcal{S}\subseteq \mathcal{N}$, $\gamma(\mathcal{S})$ can be interpreted as the profit achieved by the players in $\mathcal{S}\subseteq \mathcal{N}$. The usual goal in cooperative game is to distribute the total gain $\gamma(\mathcal{N})$ of the global coalition $\mathcal{N}$ among each player in fair and reasonable ways. Different requirements on the fairness and rationality derive different solution concepts of the cooperative game. Such as the core, the Banzhaf power index and some related concepts of approximate core. Among various solution concepts the concept of Banzhaf power index that is motivated by fairness.

For a game $\Gamma=(\mathcal{N},\gamma)$, if it is monotone, i.e., it satisfied $\gamma(\mathcal{C})\leq\gamma(\mathcal{D})$ for every pair of coalitions $\mathcal{C}, \mathcal{D}\subseteq\mathcal{N}$ such that $\mathcal{C}\subseteq\mathcal{D}$, and its characteristic function only takes value 0 and 1, i.e.,  $\gamma(\mathcal{S})\in\{0,1\}$, $\forall S\subseteq \mathcal{N}$, this game is called a simple game. In a simple game $\Gamma=(\mathcal{N},\gamma)$, the coalitions with value 1 are called to `winning', and that with value 0 are called `losing', i.e., $\forall \mathcal{S}\subseteq \mathcal{N}$, $\gamma(\mathcal{S})=1$ and $\gamma(\mathcal{S})=0$, respectively. Each coalition $\mathcal{S} \cup \{i\}$ that wins when $\mathcal{S}$ loses is called a swing for player $i\in \mathcal{N}$, because the membership of player $i$ in the coalition is crucial to the 'winning'. In fact, Banzhaf power index is to count the number of winning coalitions, when the player $\forall i\in\mathcal{N}$ joining some losing coalitions, to find the most crucial player that it can let the majority of coalitions winning.

Banzhaf power index, which yields an unique outcome in coalitional games, is proposed to measure the marginal contribution of players in the game~\cite{banzhaf1964weighted}. In simple games, the Banzhaf power index have a particular attractive interpretation: it measures the power of a player, i.e., the probability that he can influence the outcome of the game. In this paper, we use Banzhaf power index to measure the power of each feature.


\subsection{Construction of Banzhaf tree}

\begin{figure}[ht]
\begin{minipage}[b]{1\linewidth}
  \centerline{\includegraphics[height=3.5cm]{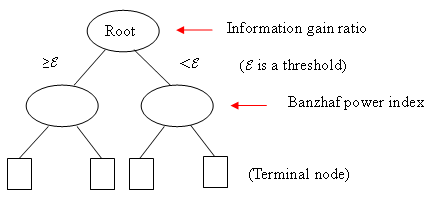}}\vspace{-0.3cm}
\end{minipage}
\caption{ A Banzhaf tree} \vspace{-0.3cm}
\label{fig:im2}
\end{figure}

Figure~\ref{fig:im2} shows the structure of a Banzhaf decision tree. For the root node, the feature is selected with information gain rate. For all the other nodes, the features are selected with the Banzhaf power index. The idea of Banzhaf decision tree are mainly motivated by game theory, especially, the cooperative game theory. We take the features of data as the players in a game, then the original tree construction problem is transformed into a cooperative `feature' game. At each node, features in the form of the coalition are selected and the best one is split.

Next, we first present the way to compute the Banzhaf power index in this work.

The original definition of Banzhaf power index is described in~\cite{banzhaf1964weighted}. Given a cooperative game $\Gamma=(\mathcal{N},\gamma)$ with $|\mathcal{N}|=n$, the Banzhaf power index of a player $i\in \mathcal{N}$ is the probability of swings for play $i$. We denote the Banzhaf power index as $\beta_i(\Gamma)$ and it is given by
\begin{eqnarray}\label{eq:banzhaf}
\beta_i(\Gamma)= \frac{1}{|2^{\mathcal{N}\setminus i}|}
\sum_{\mathcal{S}\subseteq \mathcal{N}}\Delta_i(\mathcal{S}),
\end{eqnarray}
where $\Delta_i(\mathcal{S})$ is the marginal
contribution of player $i$. i.e. $\Delta_i(\mathcal{S})=\gamma(\mathcal{S}\cup i)-\gamma(\mathcal{S})$.

Banzhaf power index measures the distribution of power among the players in cooperative games. Here, we apply it for the decision tree construction, attempting to estimate the power of each feature for each tree node. The power of each feature can be measured by averaging the contributions that it makes to each of the subset which it belongs to. Let coalition $\mathcal{K}$ be a candidate feature subset and feature $f_i(f_i\notin (\mathcal{K}))$ is to be estimated. Define the ratio $p={\mu_i(\mathcal{K})}/{\rho_i(\mathcal{K})}$ to represent the impact of feature $f_i$ on coalition $\mathcal{K}$, where $\mu_i(\mathcal{K})$ can be interpreted as the number of features that fall into interdependence relationship with the feature $f_i$, and $\rho_i(\mathcal{K})$ be the number of features in the coalition $\mathcal{K}$. Therefore, we define a threshold value $\tau$. If $p<\tau$  (commonly $\tau=1/2$), we call the coalition $\mathcal{K} \cup {f_i}$ `losing', otherwise `winning', i.e.
\begin{eqnarray}
\Delta_i(\mathcal{K} \cup {f_i})=\left\{
\begin{array}{rcl}
1      &          &  {p\geq\tau}; \\
\
0      &          &  {p < \tau }.
\end{array} \right.
\end{eqnarray}
Here, $\Delta_i(\mathcal{K} \cup {f_i}) = 1$ means that feature $f_i$ is the key to make the coalition to exhibit better performance. The threshold value 1/2 means, if more than half of the features are interdependent with $f_i$, it will join in the coalition to make it `winning'. Hence, for simplicity of the computation, we define $\Delta_i(\mathcal{S})$ in Eq.~(\ref{eq:banzhaf}) as
\begin{eqnarray}\label{eq:deltaS}
\Delta_i(\mathcal{S})=\left\{
\begin{array}{rcl}
1      &          &  {p\geq\tau}; \\
\
0      &          &  {p < \tau }.
\end{array} \right.
\end{eqnarray}
For clarity, here, we give an example to show how to compute the Banzhaf power index. Given a cooperative `feature' game $\Gamma=(\mathcal{N},\gamma)$, the feature player set $\mathcal{N}=\{f_1,f_2,f_3,f_4\}$. Suppose, currently, the goal is to calculate the Banzhaf power index of $f_4$. The total number of possible coalitions of feature subsets $\mathcal{N}\setminus f_{4}$ is 7 (except $\varnothing$), for all $\mathcal{S}\subseteq \mathcal{N}\setminus f_{4}$. Assume the winning coalitions with respect to $f_4$ are $\{f_2\}$, $\{f_2,f_3\}$, $\{f_1,f_2\}$, i.e. half of the coalitions are interdependent with feature$f_4$. Then the Banzhaf power index of $f_4$ can be computed as
\begin{eqnarray}
\beta_i(\Gamma)= \frac{1}{|2^{\mathcal{N}\setminus f_{4}}|}\sum_{\mathcal{S}\subseteq \mathcal{N}}\Delta_i(\mathcal{S})=3/7.
\end{eqnarray}
Similarly, the value of Banzhaf power index for other features can be computed as the same way. Generally, Banzhaf power index is hardly to be zero in large scale and high dimensional applications.

In order to evaluate the impact of feature $f_i$, it needs to calculate the proportion of the `winning' coalitions. That will lead to a high computational complexity, but our model only randomly selected a small group of features to compute the Banzhaf power index at each node. Hence, the computational complexity is fairly low.

To calculate the proportion of the `winning' coalitions, we use conditional mutual information of information theory to evaluate the interdependent between a single $f_j\notin\mathcal{S}\subseteq \mathcal{N}$ and the feature player $f_i\in\mathcal{S}\subseteq \mathcal{N}$. If more than half of feature players $f_i\in\mathcal{S}$ are interdependent $f_j$, then have $\Delta_j(\mathcal{S})=\gamma(\mathcal{S}\cup j)-\gamma(\mathcal{S})=1$.

In our paper, the condition mutual information is defined as the amount of the interdependent between feature player $f_j\notin\mathcal{S}$ and feature player $f_i\in\mathcal{S}$ given the feature player colation $\mathcal{S}$. It is formally
defined by
\begin{eqnarray}\label{eq:muinfo}
I(f_j;f_i|\mathcal{S}\setminus{f_i})=\sum_{x\in f_j}\sum_{y\in f_i}\sum_{z\in \mathcal{S}\setminus{f_i}}\log\frac{p(x,y|z)}{p(x|z)p(y|z)}.
\end{eqnarray}
By Eq. (\ref{eq:banzhaf}), (\ref{eq:deltaS}) and (\ref{eq:muinfo}), we can get the Banzhaf power index of each feature player for the construction of each decision tree.



\subsection{Banzhaf random forests algorithm}

Given a training data set $D_n={(X_i,Y_i)}_{i=1}^n$, it includes $n$ samples and the dimensionality of data is $M$. The procedures of the Banzhaf random forests (BRF) algorithm can be described as follows.
\begin{itemize}
\item For the construction of each Banzhaf decision tree in BRF, randomly draw $n$ samples with replacement using bootstrap and randomly select $h\ll M$ features without replacement from the training data. Base on this data set $d_n={(X_i,Y_i)}_{n\times(h+1)}\subseteq D_n={(X_i,Y_i)}_{n\times(h+1)}$, grow a recursive Banzhaf tree.
\item For the root node, the feature is selected with information gain rate. For all the other nodes, the features are selected with the Banzhaf power index. The feature associated with the corresponding node is split at the midpoint of the feature values, to generate the left and right branches.
\item If a (terminal) node has the percentage of incorrectly assigned samples less than $d$, then stop building the Banzhaf tree, where $d$ is a pre-specified number.
\item BRF predicts the labels of test data based on the votes it received from each Banzhaf tree.
\end{itemize}

Our algorithm is similar to the original algorithms of ~\cite{randfore}. Both of them used bootstrap aggregating i.e., bagging ensemble algorithm. The main difference between BRF and the algorithm of~\cite{randfore} is in how the feature associated with a node is selected. BRF uses Banzhaf power index, while Breiman's method use the Gini index. Another difference is, BRF splits each node at the midpoint of the feature values but Breiman's algorithm does not. More importantly, as shown in next section, the consistency of BRF is theoretically guaranteed, but that of Breiman's algorithm is not.

We have also tested the model of pure Banzhaf random forests, i.e. the feature of the root node is also selected via the Banzhaf power index. Their performance is generally worse than that of the BRF algorithm described as above. One reason for this result may be that the feature selected via information gain rate at the root node may present some important invariant information of data.

\subsection{Prediction}

We denote a recursive tree created in the BRF algorithm based on data $D_n={(X_i,Y_i)}_{i=1}^n$ as $g_n$, where ${(X_i,Y_i)}_{i=1}^n$ are i.i.d. pairs of random variables such that $X$ (the feature vector) takes its value in $R^d$ while $Y$ (the label) is a multiclass random variable. To make a prediction for a query point $x$, each Banzhaf decision tree computes,
\begin{eqnarray}
\zeta_{n}^{k}(x)= \frac{1}{N(A_n(x))}\sum_{(X_i,Y_i)\in A_n(x)}\delta({Y_i=k}), \nonumber
\end{eqnarray}
where $A_n(x)$ denotes the node of the tree containing $x$, and $N(A_n(x))$ is the number of points that located in $A(x)$. Then the tree prediction is the class which maximizes that:
\begin{eqnarray}
g_n(x)=\arg\max_k\{\zeta_{n}^{k}(x)\}. \nonumber
\end{eqnarray}

The forest predicts the class with the most votes from the individual trees.

\section{Consistency}

In this section, we prove the consistency of Banzhaf random forests. We denote the Banzhaf tree created by Banzhaf random forests trained on data ${(X_i,Y_i)}_{i=1}^n$ as ${g_n}$. The consistency of a sequence $\{g_n\}$ is defined as follows.

\textbf{Definition 1}  A sequence of classifier $\{g_n\}$ is consistent for a given distribution of $(X,Y)$, that is, the probability of prediction error of $g_n$ converges in probability to the Bayesian risk,
\begin{eqnarray}
L(g_n)=\mathbb{P}(g_n(X,\theta)\neq Y|D_n) \rightarrow L^\ast, \nonumber
\end{eqnarray}
as $n\rightarrow \infty$. Here, $\theta$ denotes the randomness in the tree-building algorithm, $D_n$ is the training data set and the probability in the convergence is over the random selection of $D_n$. The Bayesian risk is the probability of prediction error of the Bayesian classifier, which makes predictions by choosing the class with the highest posterior probability, $g(x)=\arg\max\limits_{k}{\mathbb{P}(Y=k|X=x)}$.

In order to reduce the complexity of the issue, we consider that multi-class classifier can be transformed to combination of several binary-class classifier. So, we need to prove the consistency of estimators of the posterior distribution of each class. A similar result was shown by Denil et al~\cite{denil2013consistency}.

\textbf{Lemma 1} Suppose we have the estimates, $\zeta_{n}^{k}(x)$, for each class posterior $\zeta^{k}(x)=\mathbb{P}(Y=k|X=x)$ and that these estimates are each consistent. The classifier
\begin{eqnarray}
g_n(x)=\arg\max_k\{\zeta_{n}^{k}(x)\} \nonumber
\end{eqnarray}
is consistent for the corresponding multi-class classification problem.

Proof. By definition, the rule
\begin{eqnarray}
g(x)=\arg\max_k\{\zeta^{k}(x)\} \nonumber
\end{eqnarray}
achieves the Bayes risk. In the case where all the $\zeta^{k}(x)$ are equal there is nothing to prove, since all choices have the same probability of error. So, suppose there is at least one $k$ such that $\zeta^{k}(x)<\zeta^{g(x)}(x)$ and define
\begin{eqnarray}
m(x)=\zeta^{g(x)}(x)-\max\limits_{k}\{\zeta^{k}(x)|\zeta^{k}(x)<\zeta^{g(x)}(x)\}   \nonumber \\
m_{n}(x)=\zeta_{n}^{g(x)}(x)-\max\limits_{k}\{\zeta_{n}^{k}(x)|\zeta^{k}(x)<\zeta^{g(x)}(x)\}  \nonumber
\end{eqnarray}

The function $m(x)\geq 0$ is the margin function which measures how much better the best choice is than the second best choice. The function $m_n(x)$ measures the margin of $g_n(x)$. If $m_n(x)>0$ then $g_n(x)$ has the same probability of error as the Bayes classifier.

The assumption above guarantees that there is some $\epsilon$ such that $m(x)>\epsilon$. Using $\mathcal{C}$ to denote the number of classes, by making $n$ large it can satisfy
\begin{eqnarray}
\mathbb{P}(|\zeta_{n}^{k}(X)-\zeta^{k}(X)|<\varepsilon/2)\geq 1-\delta/\mathcal{C} \nonumber
\end{eqnarray}
since $\zeta_{n}^{k}$ is consistent. Thus
\begin{eqnarray}
\mathbb{P}(\bigcap\limits^{\mathcal{C}}_{k=1}|\zeta_{n}^{k}(X)-\zeta^{k}(X)|<\epsilon/2)\geq 1-K+\sum\limits_{k=1}^\mathcal{C}
\mathbb{P}(|\zeta_{n}^{k}(X)-\zeta^{k}(X)|<\epsilon/2) \geq 1-\delta   \nonumber                                                       \end{eqnarray}
So with probability at least $1-\delta$ we have
\begin{eqnarray}
m_n(X)&=&\zeta_{n}^{g(X)}-\max\limits_k\{\zeta_{n}^{k}(X)|\zeta^{k}(X)<\zeta^{g(X)}(X)\}  \nonumber   \\
&\geq&(\zeta^{g(X)}-\epsilon/2)-\max\limits_k\{\zeta_{n}^{k}(X)+\epsilon/2|\zeta^{k}(X)<\zeta^{g(x)}(X)\}   \nonumber   \\
&=&\zeta^{g(X)}-\max\limits_k\{\zeta^{k}(X)|\zeta^{k}(X)<\zeta^{g(x)}(X)\}-\epsilon
>0                       \nonumber
\end{eqnarray}
Since $\delta$ ia arbitrary this means that the risk of $g_n$ converges in probability to the Bayes risk.

Lemma 1 allows us to prove the consistency of the multiclass classifier can be transformed to prove the consistency of several two class posterior estimates. i.e., Given a set of classes $\{1,...,c\}$ we can re-assign the labels using
the map $(X, Y)\mapsto (X, \mathcal I({Y=k}))$ for any $k\in \{1,...,c\}$ in order to get a two class problem where $\mathbb{P}(Y=1|X=x)$ in this new problem is equal to $\zeta^{k}(x)$ in the original multiclass problem.

Then, we are inspired by~\cite{denil2013consistency}. The following Lemma 2 allows us to focus our attention on the consistency of each of the tree estimators in the classification forests.

\textbf{Lemma 2} Assume that the sequence $\{g_n\}$ of randomized classifiers is consistent for a certain distribution of $(X,Y)$. Then the voting classifier $g_{n}^{(m)}$ obtained by taking the majority vote over $M$ (not necessarily independent) copies of $\{g_n\}$ is also consistent.

Proof. Let $g(x)$ denote the Bayes classifier. Consistency of $\{g_n\}$ is equivalent to saying that $  \mathbb{E}[L(g_n)]=\mathbb{P}(g_n(X,\theta)\neq Y) \rightarrow L^\ast$. In fact, since $\mathbb{P}(g_n(X,\theta)\neq Y|X=x)\geq \mathbb{P}(g(X)\neq Y|X=x)$ for all $x\in \mathbb{R}^D$, consistency of $\{g_n\}$ means that $\mu$-almost all $x$,
\begin{eqnarray}
\mathbb{P}(g_n(X,\theta)\neq Y|X=x)\rightarrow \mathbb{P}(g(X)\neq Y|X=x)=1-\max_k\{\zeta^{k}(x)\} \nonumber
\end{eqnarray}
Define the following indices
\begin{eqnarray}
G=\{k|\zeta^{k}(x)=\max_k\{\zeta^{k}(x)\}, B=\{k|\zeta^{k}(x)<\max_k\{\zeta^{k}(x)\} \nonumber
\end{eqnarray}
Then
\begin{eqnarray}
\mathbb{P}(g_n(X,\theta)\neq Y|X=x)=\sum_k\mathbb{P}(g_n(X,\theta)=k|X=x)\mathbb{P}(Y\neq k|X=x)  \nonumber \\
\leq(1-\max_k\{\zeta^{k}(x)\})\sum_{k\in G}\mathbb{P}(g_n(X,\theta)=k|X=x)+\sum_{k\in B}\mathbb{P}(g_n(X,\theta)=k|X=x) \nonumber
\end{eqnarray}
which means it suffices to show that $\mathbb{P}(g_{n}^{(m)}(X,\theta^{M})=k|X=x)\rightarrow 0$ for all $k\in B$. However, using $\theta^{M}$ to denotes $M$ (possible dependent) copies of $\theta$, for all $k\in B$ we have
\begin{eqnarray}
\mathbb{P}(g_{n}^{(m)}(X,\theta^{M})=k)&=&\mathbb{P}\bigg(\sum\limits^{M}_{j=1}\mathbb{I}\{g_n(x,\theta_j)=k\}>\max\limits_ {c\neq k}\sum\limits^{M}_{j=1}\mathbb{I}\{g_n(x,\theta_j)=c\}\bigg) \nonumber \\
&  & \leq \mathbb{P}(\sum\limits^{M}_{j=1}\mathbb{I}\{g_n(x,\theta_j)=k\}\geq1) \nonumber
\end{eqnarray}
By Markov's inequality,
\begin{eqnarray}
\leq \mathbb{E}[\sum\limits^{M}_{j=1}\mathbb{I}\{g_n(X,\theta_j)=k\}]   \nonumber \\
=M\mathbb{P}(g_n(X,\theta)=k)\rightarrow0. \nonumber
\end{eqnarray}

According to Lemma 2, we conclude that the consistency of Banzhaf random forests is implied by the consistency
of the trees which composed of. In addition, we use the bagging ensemble method to construct BRF. So by the Theorem 1 in~\cite{biau2008consistency}, we know that the consistency of a voting Banzhaf random forests which follows from the consistency of the base classifier. Here, Biau et al. introduce a parameter $q_n\in[0,1]$. In the bootstrap sample $D_n(\theta)$, each data pair $(X_i, Y_i)$ is present with probability $q_n$ which is independent from each other.

\textbf{Theorem 1} Let $\{g_n\}$ be a sequence of classifier that is consistency for the distribution of $(X, Y)$. Consider the Banzhaf random forests (majority voting classifiers) $g_n^{(m)}(X, \theta^m, D_n)$, using parameter $q_n$. If $nq_n\rightarrow \infty$ as $n\rightarrow \infty$ then both classifiers are consistent.

Proof. See that for Theorem 1 in~\cite{biau2008consistency}.

With Lemma 2 and Theorem 1 established, the remainder of effort goes into proving the consistency of a Banzhaf tree construction. For each tree in the Banzhaf forests is established based on the Banzhaf index. We show that if a classifier is condition consistency which consists of a small group of random variable, and uses the Banzhaf power index to sampling for this sample process for this random variable generates acceptable sequences with probability 1, then the resulting classifier is unconditionally consistent.

\textbf{Theorem 2} Suppose $\{g_n\}$ is a sequence of classifiers whose probability of error converges conditionally in probability to the Bayes risk $L^\ast$ for a specified distribution on $(X,Y)$, i.e.
\begin{eqnarray}
\mathbb{P}(g_n(X,\theta,I)\neq Y|I)\rightarrow L^\ast,   \nonumber
\end{eqnarray}
for all $I\in \mathcal{I}$, $I$ is a random sequence produced by Banzhaf power index, and that $v$ is a distribution on $I$. If $v(\mathcal{I})=1$ which means produce acceptable sequence with probability value is 1, then the probability of error converges unconditionally in probability, i.e.\begin{eqnarray}\mathbb{P}(g_n(X,\theta,I)\neq Y)\rightarrow L^\ast,\nonumber \end{eqnarray} $\{g_n\}$ is consistent for the specified distribution.

Proof. The sequence in question is uniformly integrable, so it is sufficient to show that $\mathbb{E}[\mathbb{P}{(g_n(X,\theta,I)\neq Y|I)}]\rightarrow L^\ast$ implies the result, where the expectation is taken over the random selection of training set and $I$ is the specific structure of the tree, $\{g_n\}$. We can write
\begin{eqnarray}
\mathbb{P}(g_n(X,\theta,I)\neq Y)&=&\mathbb{E}[\mathbb{P}(g_n(X,\theta,I)\neq Y|I)] \nonumber \\
&=&\int_\mathcal{I}\mathbb{P}(g_n(X,\theta,I)\neq Y|I)v(I)+\int_{\mathcal{I}^c}\mathbb{P}(g_n(X,\theta,I)\nonumber \neq Y|I)v(I) \end{eqnarray}
By assumption $v(\mathcal{I}^c)=0$ then we have\begin{eqnarray}\lim_{n \to \infty} \mathbb{P}(g_n(X,\theta,I)\neq Y)=\lim_{n \to \infty}\int_\mathcal{I}\mathbb{P}(g_n(X,\theta,I)\neq Y|I)v(I) \nonumber  \end{eqnarray}
Since probabilities are bounded in the interval $[0,1]$, the dominated convergence theorem allows us to exchange the integral and the limit,\begin{eqnarray}=\int_\mathcal{I}\lim_{n \to \infty}\mathbb{P}(g_n(X,\theta,I)\neq Y|I)v(I) \nonumber  \end{eqnarray}and by assumption the conditional risk converges to the Bayes risk for all $I\in \mathcal{I}$, so\begin{eqnarray}=L^\ast\int_\mathcal{I}v(I) \nonumber
=L^\ast\end{eqnarray} which is the desired result.

In fact, let the Banzhaf power index $\eta(f_i)$ is equal to the income distribution function $\gamma(f_i)$ in a tree construction game $\Gamma=(\mathcal{N},\gamma)$ ,i.e., $\eta(f_i)=\gamma(f_i)$. Because we chose the maximize Banzhaf power index for each node of each tree. We can obtain a acceptable random variable sequence that all with the maximize Banzhaf power index. By $\eta(f_i)=\gamma(f_i)$, these random variable sequence cooperative can obtain the best result. So it is sufficient to show that the Banzhaf tree is consistent conditioned on such a sequence.

In conclusion, we proved the consistency of our tree construct by the Theorem 2. Because the Theorem 1 is established, we can achieve the consistency of Banzhaf random forests.


\section{Experiments}

To evaluate the proposed algorithm, BRF, we tested it on several data sets from the UCI machine learning repository, including iris, wine, ecoli, thyroid, soybean, shuttle, dermatology, sonar and musk2. We compare it with Breiman's random forests~\cite{randfore} and the model proposed in~\cite{biau2012analysis}. We implemented Breiman's random forest with C4.5 as it generally performs well on classification problems. As mentioned above, the model proposed in~\cite{biau2012analysis} is consistent. For comparison, we also listed the classification results yielded by k-nearest neighbor classifier (KNNs) and support vector machines (SVM).

Table~\ref{tab:t1} shows the specific information of the used UCI data sets.

\begin{table}[ht]
  \small
  \centering
  \begin{tabular}{c |c |c |c }
  \hline
 \textbf  {Datasets} & No.examples& No.features & No.classes \\ \hline
     soybean         &47          &35            &4\\
     iris            &150         &4             &3\\
     wine              &178        &13       &3\\
     sonar             &208        &20       &2\\
     thyroid           &215        &5        &3 \\
     ecoli             &357        &7        &8 \\
     dermatology       &366        &34       &6 \\
     musk2             &6598       &166      &2 \\
     shuttle           &14516      &9        &7 \\ \hline
  \end{tabular}
  \setlength\abovecaptionskip{10pt}
  \caption{Summary of the used UCI data sets.}
  \label{tab:t1}
\end{table}

\subsection{Effect of the number of trees in BRF}

\begin{figure}[ht]
\begin{minipage}[b]{1\linewidth}
  \centerline{\includegraphics[height=4.5cm]{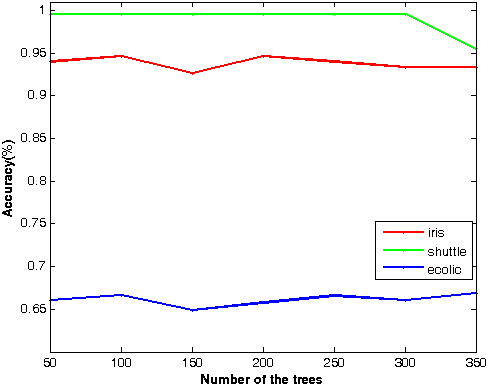}}
\end{minipage}
\caption{Effect of the number of trees in BRF.}
\label{fig:im1}
\end{figure}

To evaluate the effect of the number of trees in BRF, we conducted experiments on three data sets: iris, ecoli and shuttle. Fig.~\ref{fig:im1} shows the obtained classification accuracy against the number of trees in BRF. We can see that, BRF is basically robust with the number of trees. Particularly, when the number of trees equals to 100, BRF performs slightly better than other values.

\subsection{Comparison on running efficiency}

To test the running speed of BRF, we performed experiments on seven data sets: iris, wine, ecoli, soybean, thyroid, dermatology and shuttle. We compared it with the model of~\cite{randfore} and that of~\cite{biau2012analysis}. From Table~\ref{tab:t3}, we can see that, the running of BRF is slower than the model of~\cite{randfore}. This is mainly because calculation of the Banzhaf power index needs some time when constructing the trees. However, BRF is more efficient than the model of~\cite{biau2012analysis}, which is a state-of-the-art consistent random forests model.
\begin{table}[ht]
  \small
  \centering
  \begin{tabular}{c  |c  |c  |c  }
  \hline
  \textbf {Datasets}   & Breiman01 & Biau12 & BRF \\\hline
    iris           & 1.321   & 3.107 & 1.654\\
    wine           & 5.401   & 16.781 & 9.134\\
    ecoli          & 5.729   & 17.438 & 8.778\\
    soybean        & 0.673   &5.761   &2.297\\
    thyroid        & 2.857  & 4.856 & 3.168\\
    dermatology    & 2.463  & 71.201 & 11.023\\
    shuttle        & 49.71  & 39600.63& 80.660\\\hline
  \end{tabular}
  \setlength\abovecaptionskip{10pt}
  \caption{Running time of two compared models and BRF on seven UCI data sets (the unit is second).}
  \label{tab:t3}

\end{table}

\subsection{Classification results}

To evaluate BRF on multi-class classification problems, we compared it with KNNs, SVMs, the model of~\cite{randfore}, and the model of~\cite{biau2012analysis}. Nine UCI data sets were used. They are iris, wine, ecoli, thyroid, soybean, shuttle, dermatology, sonar and musk2. For all these data sets, we used 5-fold cross validation to test the models. The average classification accuracies are reported. For the model of~\cite{randfore} and BRF, we used the same number of trees in the random features. Following Breiman's suggestion for classification problems~\cite{randfore}, we set the number of trees to $round(\log2(h)+1)$, where $h$ is the dimensionality of features. To be fair, we set up the same termination conditions for all the random forests models, i.e. the percentage of incorrectly assigned samples at the termination node should be no greater than the number of classes on a data set. For KNNs and SVMs, we selected the parameter with 5-fold cross validation.

Table~\ref{tab:t2} shows the results obtained by the compared models and BRF. We can see that BRF performs slightly better than KNNs, SVMs and the model of~\cite{randfore}, and consistently better than the model of~\cite{biau2012analysis}. This demonstrates that using interdependent features to construct the randomized trees can lead to better results than using independent features in random forests.

\begin{table}[ht]
  \small
  \centering
  \begin{tabular}{c |c |c |c |c |c }
  \hline
  \textbf {Datasets}  & KNN    & SVM    & Breiman01 & Biau12 & BRF \\\hline
  soybean            & \textbf{1.0000} & \textbf{1.0000} & \textbf{1.0000}   & 0.5717 & \textbf{1.0000}       \\
  iris               & 0.9467 & \textbf{0.9867} & 0.9467   & 0.8353 & 0.9467        \\
  wine               & 0.9423 & 0.6782 & 0.9599   & 0.5580 & \textbf{0.9717}      \\
  sonar              & 0.5908 & 0.6583 & 0.7032   & 0.5819 & \textbf{0.7120}        \\
  thyroid            & 0.9395 & 0.9023 & \textbf{0.9488}   & 0.8000 & 0.9395       \\
  ecoli              & 0.8356 & \textbf{0.8431} & 0.5958   & 0.4286 & 0.6665       \\
  dermatology        & 0.9656 & 0.9540 & 0.9589   & 0.4397 & \textbf{0.9677}       \\
  musk2              & 0.7227 & 0.8508 & 0.8509   & 0.6542 & \textbf{0.8710}       \\
  shuttle            & 0.9951 & 0.9752 & 0.9957   & 0.8256 & \textbf{0.9957} \\\hline
  \end{tabular}
  \setlength\abovecaptionskip{10pt}
  \caption{Classification accuracy obtained by the compared models and BRF on the UCI data sets.}
  \label{tab:t2}
\end{table}

\section{Conclusion}

In this paper, we propose a novel random forests model called Banzhaf random forests (BRF) based on the concepts of the cooperative game theory. It's consistency is proved, which takes a step towards narrowing the gap between the theory and practice of random forest. This work is probably the first one that apply the cooperative game theory to random forests, and we have tested and verified the feasibility of the idea. Experiments on UCI data sets show that BRF not only slightly outperforms state-of-the-art classifiers, including KNNs, SVMs and the random forests model by Breiman~\cite{randfore}, but much more efficient than existing consistent random forests.

\section*{Acknowledgment}

This research was supported by the National Natural Science Foundation of China (NSFC) under Grant no. 61271405 and 61403353, and the Fundamental Research Funds for the Central Universities of China.


%
%

\end{document}